\documentclass[letterpaper, 10 pt, conference]{ieeeconf}  % Comment this line out if you need a4paper

\IEEEoverridecommandlockouts                              % This command is only needed if 
\usepackage{amsmath,amssymb,amsfonts}
\usepackage{balance}
\usepackage{cite}
\usepackage{algorithmic}
\usepackage{graphicx}
\usepackage{textcomp}
\usepackage{xcolor}
\usepackage{csquotes}
\usepackage{siunitx}
\usepackage{subcaption}
\usepackage{gensymb}
\usepackage{url}
\usepackage{multirow}
\usepackage{colortbl}
\usepackage{hyperref}
\usepackage{cleveref}
\usepackage{wasysym}

\title{\LARGE \bf
Self-Mixing Laser Interferometry for Robotic Tactile Sensing
}

\author{Remko Proesmans$^{1}$, Ward Goossens$^{1}$, Lowiek Van den Stockt$^{1}$, Lowie Christiaen$^{1}$ and Francis wyffels$^{1}$% <-this % stops a space
\thanks{*This work was supported by the Research Foundation Flanders (FWO) under grant agreement no. 1S15925N and by the euROBIN Project (EU grant number 101070596). }% <-this % stops a space
\thanks{$^{1}$Remko Proesmans, Ward Goossens, Lowiek Van den Stockt, Lowie Christiaen and Francis wyffels are with the AI and Robotics Lab (IDLab-AIRO), Ghent University---imec, Ghent, Belgium
        {\tt\small remko.proesmans@ugent.be}}%
}

\def\BibTeX{{\rm B\kern-.05em{\sc i\kern-.025em b}\kern-.08em
    T\kern-.1667em\lower.7ex\hbox{E}\kern-.125emX}}
\begin{document}

\maketitle
\thispagestyle{empty}
\pagestyle{empty}

%%%%%%%%%%%%%%%%%%%%%%%%%%%%%%%%%%%%%%%%%%%%%%%%%%%%%%%%%%%%%%%%%%%%%%%%%%%%%%%%
\begin{abstract}

Self-mixing interferometry (SMI) has been lauded for its sensitivity in detecting microvibrations, while requiring no physical contact with its target.
In robotics, microvibrations have traditionally been interpreted as a marker for object slip, and recently as a salient indicator of extrinsic contact.
We present the first-ever robotic fingertip making use of SMI for slip and extrinsic contact sensing.
The design is validated through measurement of controlled vibration sources, both before and after encasing the readout circuit in its fingertip package.
Then, the SMI fingertip is compared to acoustic sensing through four experiments.
The results are distilled into a technology decision map.
SMI was found to be more sensitive to subtle slip events and significantly more resilient against ambient noise. 
We conclude that the integration of SMI in robotic fingertips offers a new, promising branch of tactile sensing in robotics.
Design and data files are available at \href{https://github.com/RemkoPr/icra2025-SMI-tactile-sensing}{https://github.com/RemkoPr/icra2025-SMI-tactile-sensing}.

\end{abstract}

%%%%%%%%%%%%%%%%%%%%%%%%%%%%%%%%%%%%%%%%%%%%%%%%%%%%%%%%%%%%%%%%%%%%%%%%%%%%%%%%
\section{INTRODUCTION}

\begin{figure}
\centering
\includegraphics[width=0.65\linewidth]{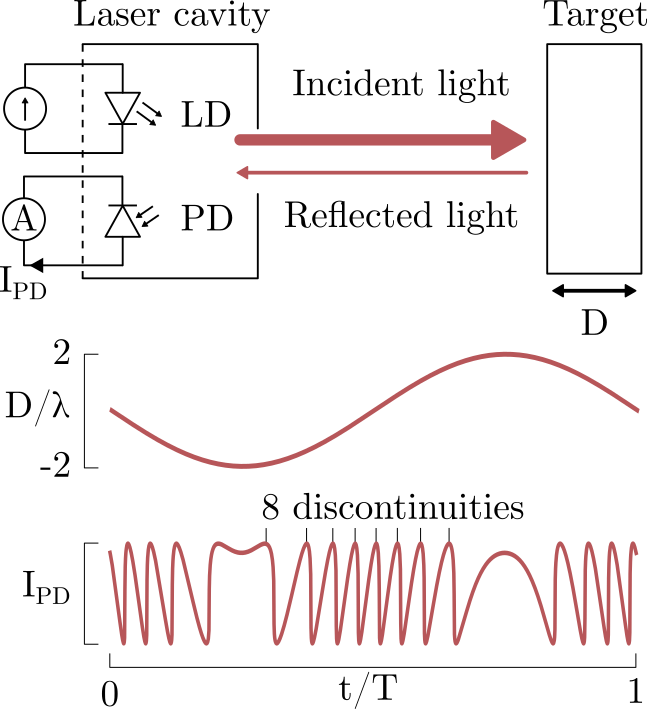}
\caption{Self-mixing interferometry. Every time the target displacement D changes by $\lambda$/2, a discontinuity or fringe appears in the SMI signal. In the example, a D of 4$\lambda$ peak-to-peak leads to 8 discontinuities in I$_{\mathrm{PD}}$.}
\label{fig:smi_principle}
\end{figure}

Self-mixing is a phenomenon where light emitted from a laser diode (LD) is reflected back into the laser cavity and interferes with the resonating light in the cavity~\cite{yu2016}.
By monitoring the light power in the cavity with a photodiode (PD), the interference pattern can be measured.
This technique is called self-mixing interferometry (SMI).
Fig.~\ref{fig:smi_principle} shows an SMI signal simulated based on \cite{plantier2005}.
The characteristic shape of the PD monitor current I$_{\text{PD}}$ is explained as follows: the distance between the laser and the target determines the phase difference between the resonating light in the cavity and the reflected light.
The phase difference determines the light power in the cavity when both beams interfere. 
When the target moves by half a laser wavelength $\lambda/2$, the optical path difference for the reflected light is $\lambda$ and the phase difference at the PD is the same.
Hence, a fringe occurs for each $\lambda/2$ that the target travels.
%The width of each fringe is determined by the linear speed of the target, which varies sinusoidally in Fig~\ref{fig:smi_principle}.
%The amount of fringes per displacement period of the target is determined by the amplitude of the target displacement: a fringe occurs for each $\lambda$ that the target travels.

In the previous decade, SMI has been widely applied across various fields, including the measurement of mechanical transfer functions of diverse objects such as car doors and micro-electromechanical systems (MEMS)~\cite{donati2018}.
Additionally, SMI has proven valuable in biomedical applications, enabling the non-contact measurement of physiological signals like blood pulsation and respiratory sounds~\cite{donati2014}.
Overall, SMI has been lauded for its highly accurate vibrometric capabilities without the need for mechanical contact with the target~\cite{donati2018, donati2014}.

Laser sensing has found very little application in robotics: \cite{maldonado2012} uses a laser navigation sensor designed for computer mice, and \cite{morita2015} uses a laser Doppler velocimeter (LDV) for slip detection.
The former does not exploit any interferometric properties of the laser, the latter requires a complex integrated structure of lenses and mirrors to do so.
SMI, compared to other interferometric techniques like LDV, has the benefit of a minimal part-count and of being self-aligned~\cite{yu2016, donati2014}.

In robotics, the microvibrations that SMI is so attuned to have been used as markers for slip during robotic manipulation.
Such microvibrations have traditionally been captured using accelerometers, piezoelectric elements and microphones~\cite{kyberd2023, slipreview2020}.
Recently, microphones have also been successfully applied to detect vibrations caused by extrinsic contact during robot manipulation tasks.
This acoustic information was found to ease the learning of
contact-rich robot manipulation skills~\cite{liu2024maniwav}.
Microphones share the characteristic property of SMI systems that no mechanical contact with the measurement target is required.
In the case of tactile sensing, this means that the sensor can be located behind the finger surface without getting in the way of structural or other electronic parts.
A second analogy is that the microvibrations caused by slip, and potentially measurable with SMI, are nothing but mechanically coupled sound waves.
Hence, microphones and SMI both essentially measure sound.
%Slip detection approaches: normal/shear force ration, movement, incipient slip, end with microvibrations. \cite{slipreview2020} 
We conclude that acoustic sensing is a topical baseline to compare SMI against.

In summary, SMI has been highly effective in vibrometry and represents new territory in robotic tactile sensing. 
We identified two important analogies between SMI and acoustic sensing and can therefore use acoustic sensing as a baseline.
The similarities are: (1) they both measure microvibrations/sound, and (2), they both measure without mechanical contact.
Our contributions are:
\begin{itemize}
    \item Design and validation of a novel robotic fingertip with SMI sensing.
    \item Experimental evaluation of SMI sensing in the context of (1) slip detection and (2) measuring extrinsic contact for robot learning, through direct comparison with an embedded microphone.
\end{itemize}

\section{Fingertip Design}

This section presents the electrical and mechanical design of both the SMI and the microphone fingertip. 
The complete design files are available on GitHub\footnote{https://github.com/RemkoPr/icra2025-SMI-tactile-sensing}.

\subsection{Circuit Design}

\subsubsection{SMI sensor} 
\begin{figure}
    \centering
    \vspace{2mm}
    \includegraphics[width=\linewidth]{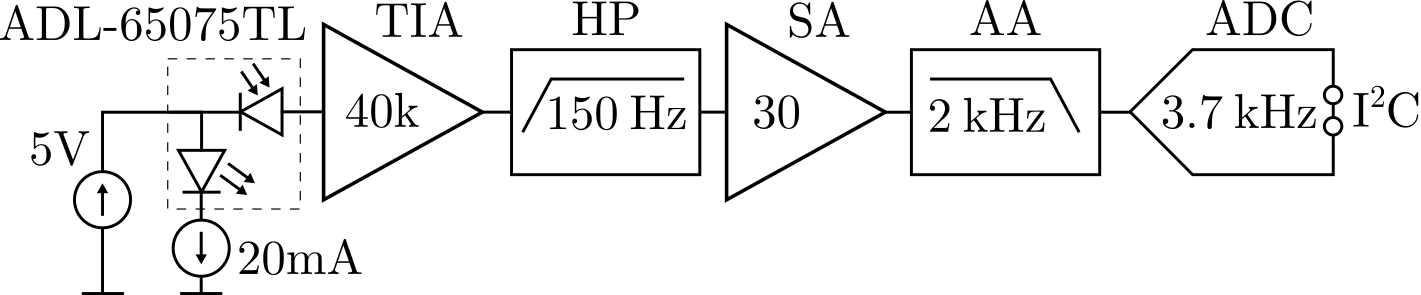} 
    \caption{The readout circuit consists of a transimpedance amplifier (TIA), a high-pass (HP) filter, a signal amplifier (SA), a low-pass anti-aliasing (AA) filter, and an analog-to-digital converter (ADC).}
    \label{fig:circuit}
\end{figure}

A schematic representation of the SMI circuit is shown in Fig.~\ref{fig:circuit}.
The laser is an ADL-65075TL ($\lambda$\,=\,650\,nm) with three leads, meaning the PD and LD have one terminal in common.
A regulated \SI{5}{V} source provides a constant voltage bias for the PD.
The LD current is fixed by a \SI{20}{mA} current regulator.
The monitor current of the PD is transduced to a voltage and amplified by a transimpedance amplifier (TIA).
This amplifies both the DC and AC components of the PD current with a transimpedance gain of 40\,000.
However, the DC component is irrelevant to the interferometric signal.
Hence, a first-order high-pass (HP) filter stage rejects the DC component.
The cutoff frequency is set to 150\,Hz, effectively attenuating both the 50\,Hz mains noise and the 100\,Hz harmonics typically generated by full-wave rectifiers in power supplies.
The AC component is further amplified by a signal amplifier (SA) with a gain of 30.
A second order Sallen-Key anti-aliasing (AA) filter removes any out-of-band noise before sampling the signal with an analog-to-digital converter (ADC).
The cutoff frequency of the AA filter is set to 2\,kHz, following \cite{su2015} and \cite{romeo2020} that have respectively identified 60-700\,Hz and 200-1000\,Hz as relevant frequency ranges for slip detection. 

\subsubsection{Microphone} The microphone used is a bottom-port ICS\nobreakdash-43434 with I2S communication interface on a custom breakout PCB.

\subsection{Mechanical Design} \label{ss:mech}

\begin{figure}[tb]
    \centering
    \begin{minipage}[b]{0.3\textwidth}
        \centering
        \includegraphics[width=0.8\textwidth]{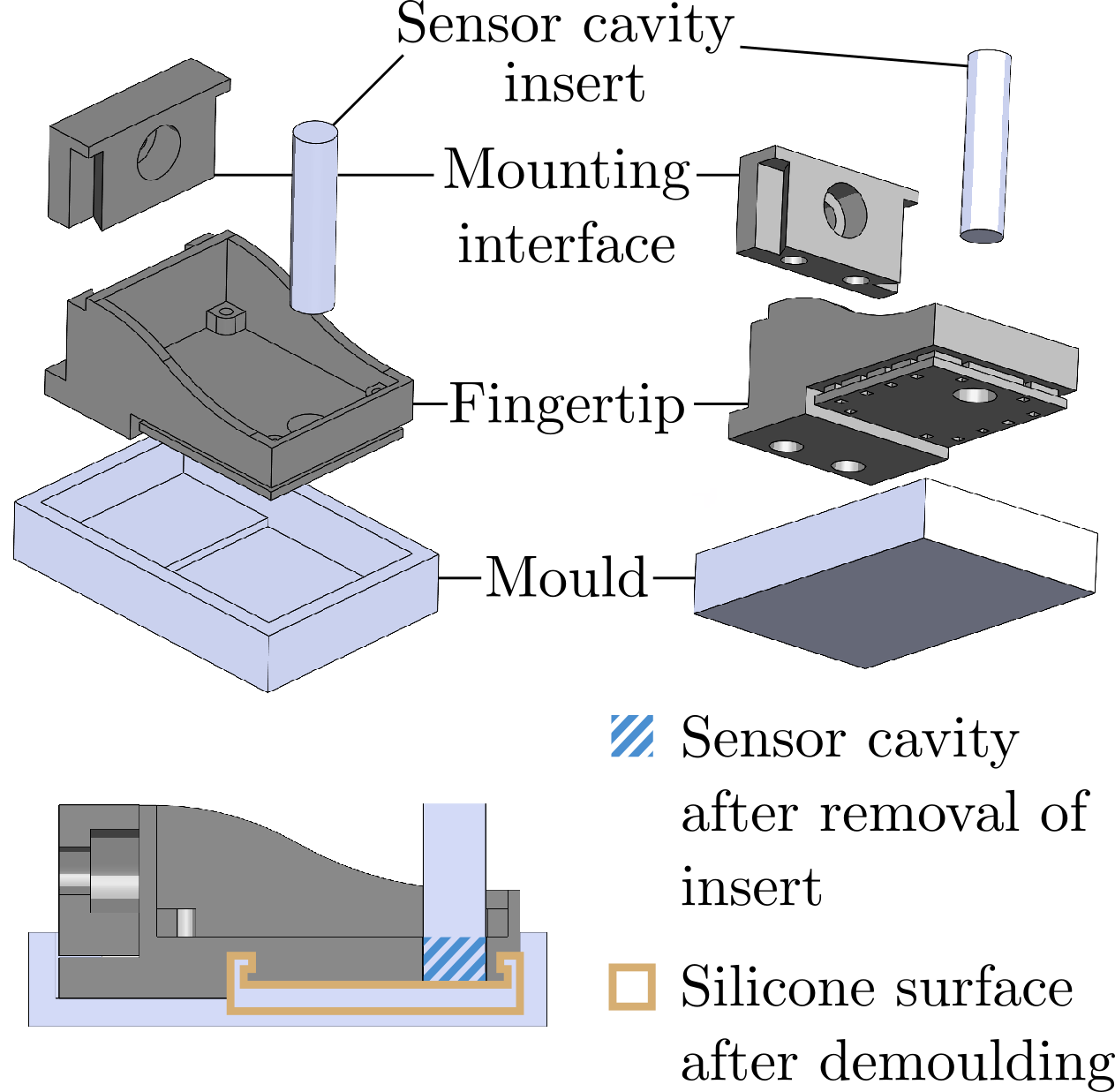} % Top left image
        \subcaption{Mould for silicone pour. }
        \label{fig:mold}
        \vspace{2mm} % Space between the two left images
        \includegraphics[width=0.7\textwidth]{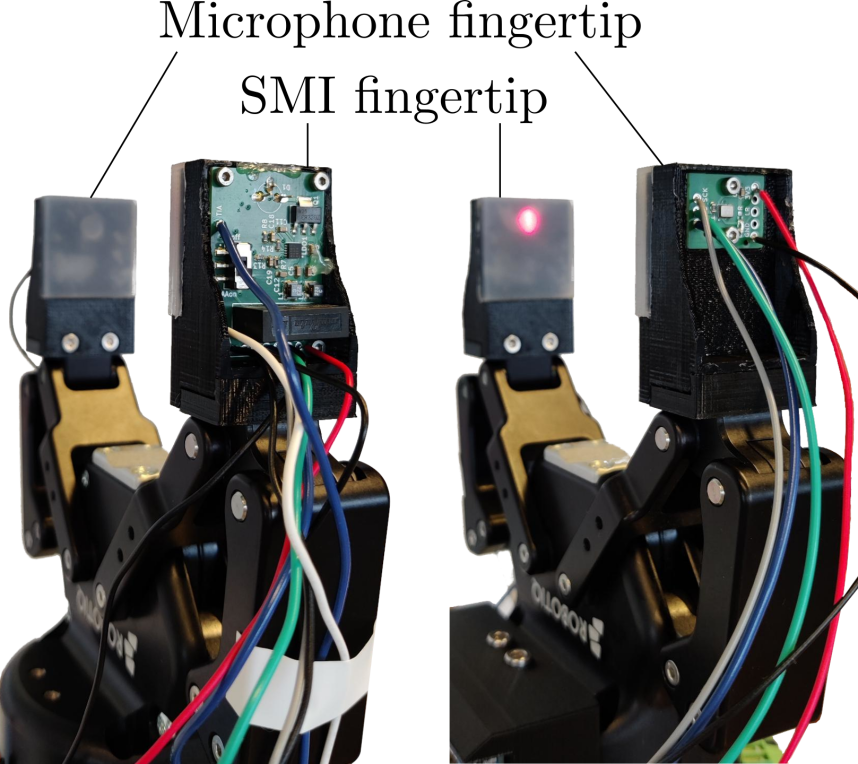} % Bottom left image
        \subcaption{Front and back views of the fingertips mounted to a Robotiq~2F\nobreakdash-85.}
        \label{fig:fingertips}
    \end{minipage}% 
    \begin{minipage}[b]{0.15\textwidth}
        \centering
        \includegraphics[height=8cm]{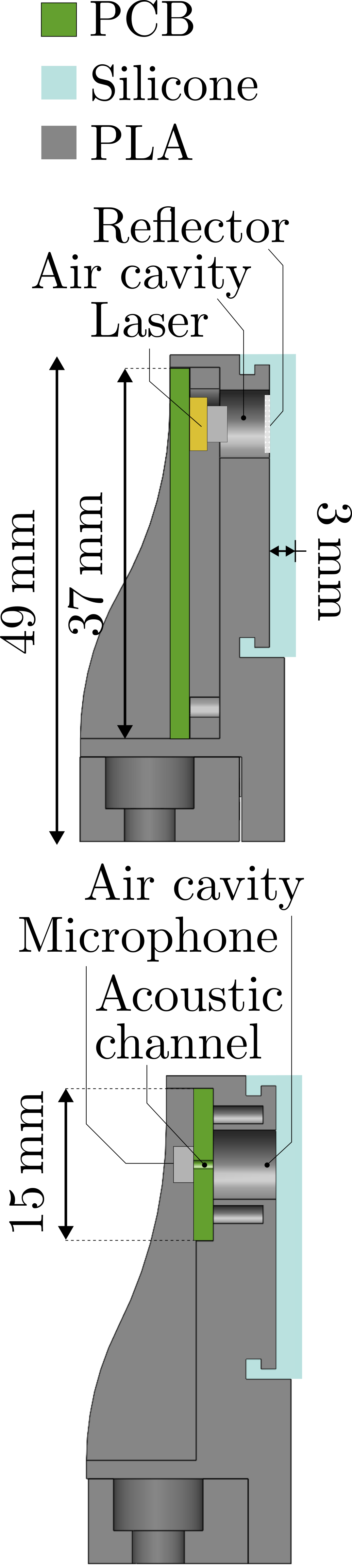} % Right image spanning both heights
        \subcaption{Section views.}
        \label{fig:section_view}
        \vspace{3.4mm}
    \end{minipage}\hspace{5mm}
    \caption{Mechanical design of the fingertips. Fingertip dimensions are equal, apart from the PCB slot.}
    \label{fig:mechanical}
\end{figure}

The fingertips consist of 3D printed polylactic acid parts, as well as a 3\,mm thick silicone contact surface.
The mould for the silicone pour is shown in Fig.~\ref{fig:mold}, the silicone used is a 50-50 mix of Silicone Addition Colorless 5 and Silicone Addition Colorless 50 by Silicones and More.
The mounting interface, indicated in Fig.~\ref{fig:mold}, is designed for mounting the fingertips to a Robotiq~2F\nobreakdash-85 gripper.
The fingertip body slides onto the mount via a dovetail sliding joint and is fixed with two M3 bolts.
The result is shown in Fig~\ref{fig:fingertips}.
Lastly, Fig.~\ref{fig:section_view} shows section views of both fingertips.
The laser is facing a piece of retroreflective tape stuck to the back of the silicone contact surface.
Since the laser is so close to its target, a focusing lens is not required.
%Arguably, SMI would be more effective when pointed directly at the target through a hole in the silicone layer, similarly to~\cite{morita2015}, but this would make the design less robust and would obstruct potential integration of other sensing modalities on the fingertip surface.
The acoustic channel of the bottom-port microphone leads to an enclosed air cavity right behind the silicone.
This way, the microphone can pick up vibrations propagating through the finger contact surface, similarly to the laser. 

%\begin{figure}[tpb]
%\centering
%\vspace{2.7mm}
%\begin{subfigure}[t]{0.35\textwidth}
%    \centering
%    \includegraphics[height=4.5cm]{img/mold.png} 
%    \caption{Mould for silicone pour. }
%    \label{fig:mold}
%\end{subfigure}
%\begin{subfigure}[t]{0.10\textwidth}
%    \centering
%    \includegraphics[height=4.5cm]{img/section_view.png} 
%    \caption{Section views. }
%    \label{fig:section_view}
%\end{subfigure}
%\vspace{2mm}%

%\begin{subfigure}[t]{0.45\textwidth}
%    \centering
%    \includegraphics[width=0.5\linewidth]{img/fingertips.png} 
%    \caption{Front and back views of the manufactured fingertips.}
%    \label{fig:fingertips}
%\end{subfigure}
%\caption{Mechanical design of the fingertips.}
%\label{fig:mechanical}
%\end{figure}

\section{Validation} \label{s:validation}
The SMI circuit is validated through two experiments.
The first will show the proper functioning of the circuit as an interferometer.
The second will show the fingertip working as a vibrometer.

\subsection{Interferometry}
\begin{figure}[tpb]
\centering
%\vspace{2.7mm}
\begin{subfigure}[t]{0.13\textwidth}
    \centering
    \includegraphics[height=4cm]{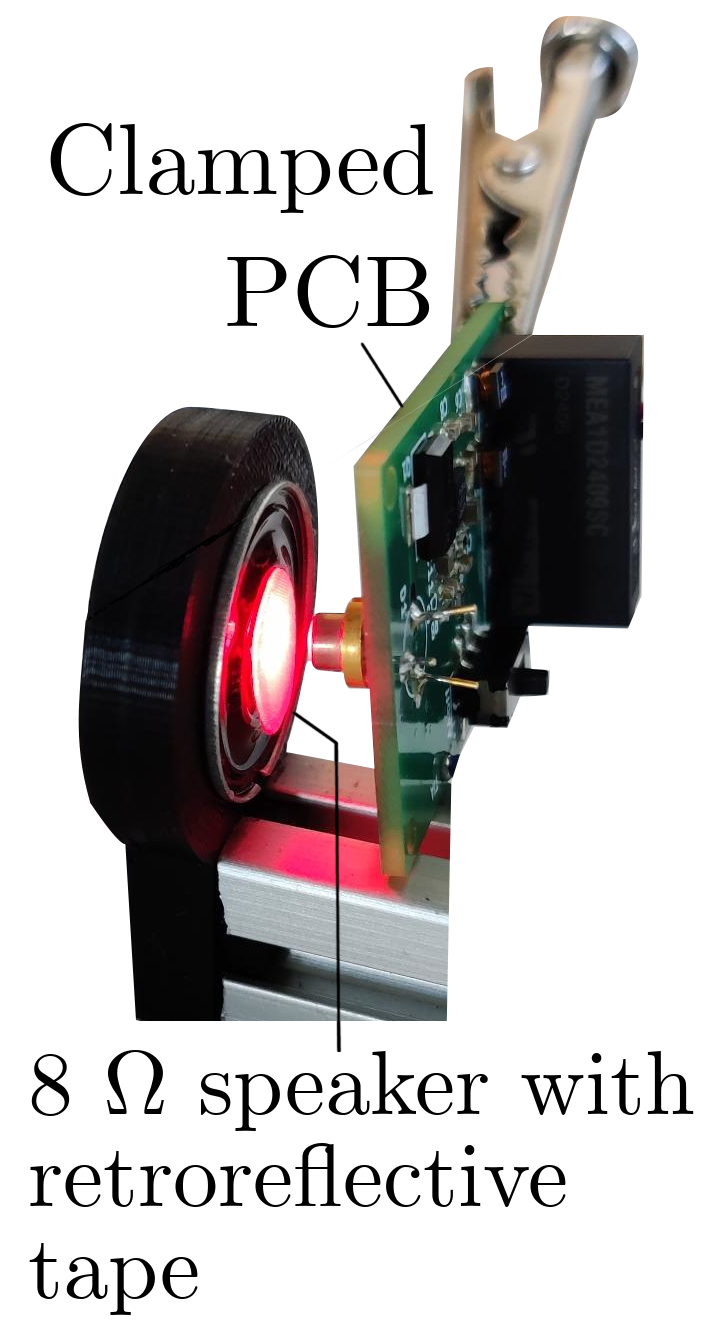}
    \caption{Setup.}
    \label{fig:smi_setup}
\end{subfigure}\hspace{1mm}
\begin{subfigure}[t]{0.34\textwidth}
    \includegraphics[width=\linewidth]{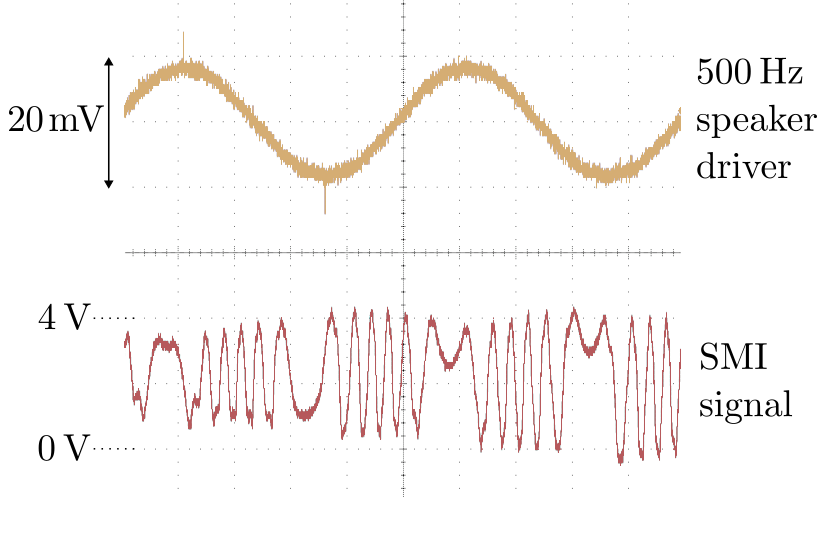} 
    \caption{Oscilloscope measurement of the SA output.}
    \label{fig:smi_measurement_annotated}
\end{subfigure}
\caption{Measurement of the SMI signal when the laser is pointed towards an 8\,$\Omega$ speaker driven at 500\,Hz.}
\label{fig:smi}
\end{figure}

The finished PCB is clamped with the laser facing a \diameter 2\,cm \SI{8}{\text{$\Omega$}} speaker, see Fig.~\ref{fig:smi_setup}.
A piece of retroreflective tape was attached to the speaker to increase the amount of reflected light.
The speaker is driven with a 20\,mV sinusoid at 500\,Hz.
The output of the signal amplifier is measured using a MSO~4032 oscilloscope set to 8x averaging, see Fig.~\ref{fig:smi_measurement_annotated}. 
The measured signal clearly resembles the simulated SMI signal from Fig.~\ref{fig:smi_principle}, and thus confirms the proper functioning of the circuit.
For each half-period, we count 6 discontinuities, meaning the speaker has moved by about 3$\lambda$, which makes for 1.95\,\textmu m with a $\lambda$ of 650\,nm.

\subsection{Vibrometry}
\begin{figure}[tpb]
\centering
%\vspace{2.7mm}
\begin{subfigure}[t]{0.1\textwidth}
    \centering
    \includegraphics[height=4.2cm]{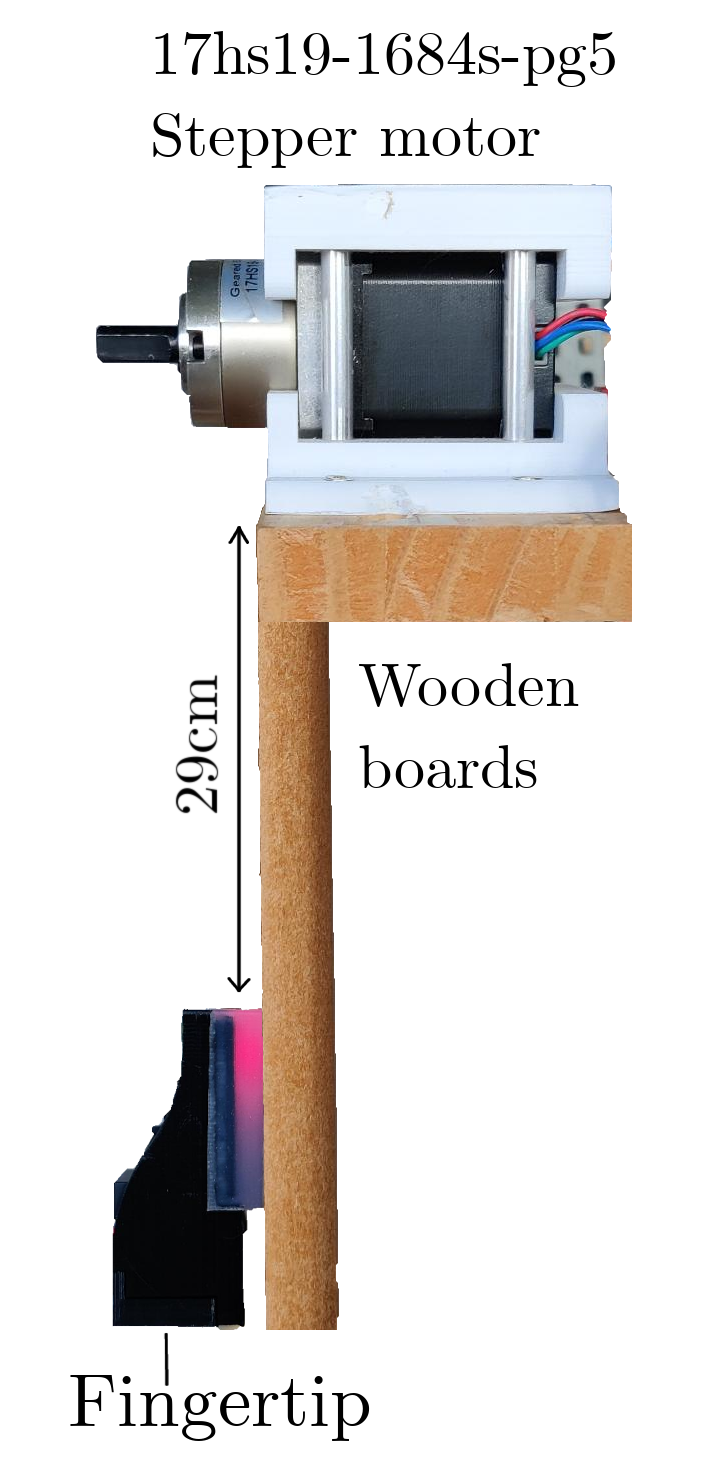}
    \caption{Setup.}
    \label{fig:vibrometer_setup}
\end{subfigure}\hspace{1mm}
\begin{subfigure}[t]{0.37\textwidth}
    \centering
    \includegraphics[width=\linewidth]{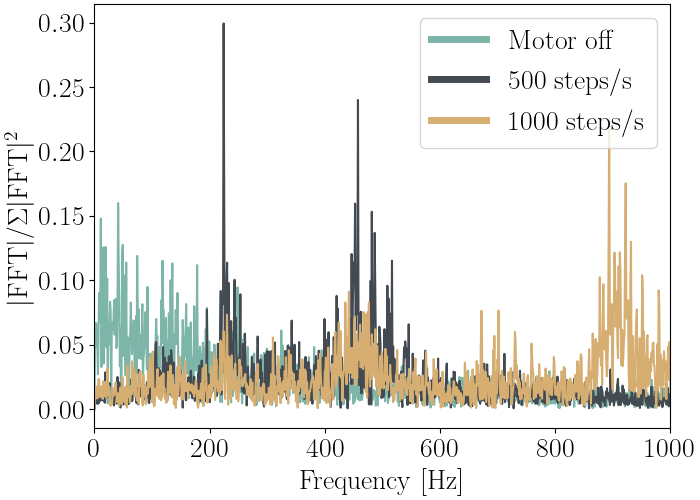} 
    \caption{Normalised fourier spectrum of a 1 second window of the ADC output.}
    \label{fig:vibrometer_plot}
\end{subfigure}
\caption{Vibrometer experiment. The SMI fingertip is pushed against a wooden board that is mechanically connected to a turning stepper motor. Setup is edited not to scale.}
\label{fig:vibrometer}
\end{figure}

As Fig.~\ref{fig:smi_principle} and \ref{fig:smi_measurement_annotated} show, an SMI signal is periodic with the fundamental frequency of the exciting wave.
This allows to use SMI for vibrometry.
However, it remains to be confirmed that the mechanical fingertip structure described in section~\ref{ss:mech} adequately captures vibrations.
To test this, the SMI fingertip is pushed against a wooden board that is mechanically connected to a turning stepper motor, see Fig.~\ref{fig:vibrometer_setup}. 
The motor is driven at 500\,steps/s and\,1000 steps/s, and normalised Fourier spectra of the SMI output are plotted in Fig.~\ref{fig:vibrometer_plot}.
We note some distortion: the spectra show peaks 50-100\,Hz below the driving signal, as well as at half the driving frequency.
That being said, the SMI sensor produces clear, reproducible signals of vibrations propagated through 30\,cm of wood.
We conclude that the distortions introduced by the mechanical fingertip design are inconsequential for the purposes of slip and extrinsic contact detection.

\section{Robot Manipulation Experiments} \label{s:experiments}

In this section, four experiments are performed to compare the sensitivity of SMI sensing with that of acoustic sensing.
In the first two experiments, the sensor outputs are compared when an object slips between the fingers, and the effect of slip speed is evaluated.
In the other two, extrinsic contact is detected with variations in the ambient noise level (ANL).

\subsection{Hardware setup \& Methodology}
Both fingertips are mounted to a Robotiq~2F\nobreakdash-85 gripper on a UR5e robotic arm.
The SMI sensor can be read directly by an Arduino Uno or by our Halberd coupling~\cite{halberd} through a 3.3\,V--5\,V level shifter.
The microphone is read in parallel by an Arduino MKR\,1000.

In this paper, we limit the analysis to the time domain.
The output of the SMI circuit in Fig.~\ref{fig:circuit} is in volts, the output of the ICS\nobreakdash-43434 microphone is in amplitude units that can be related to sound pressure level.
To compare the sensor output amplitudes A, we will normalise them by the square root of their noise floor power P$_{\text{noise}}$, i.e. the signal power at rest:
\begin{equation}
    \sqrt{\text{P}_{\text{noise}}}=\sqrt{\frac{1}{N}\sum_{i=0}^{N-1} |x_{\text{noise}, i}-\bar{x}_\text{noise}|^2}
\end{equation}
with $N$ the number of measured samples $x_{\text{noise}, i}$, $\bar{x}_\text{noise}$ being the mean of all $x_{\text{noise}, i}$.
The signal power (e.g. during a slip event) divided by P$_{\text{noise}}$ is the signal-to-noise ratio (SNR).
The SMI circuit was found to exhibit fluctuations in the noise floor.
For this reason, $\text{P}_{\text{noise}}$ is always calculated over time intervals of high noise to obtain the worst-case noise power estimate.
Hence, the stationary noise of the microphone will often be seen to rise above the noise of the laser in the graphs of the upcoming subsections.

\subsection{Cable slip}

\begin{figure}[tpb]
\centering
\begin{subfigure}[t]{0.48\textwidth}
    \centering
    \vspace{2mm}
    \includegraphics[height=3cm]{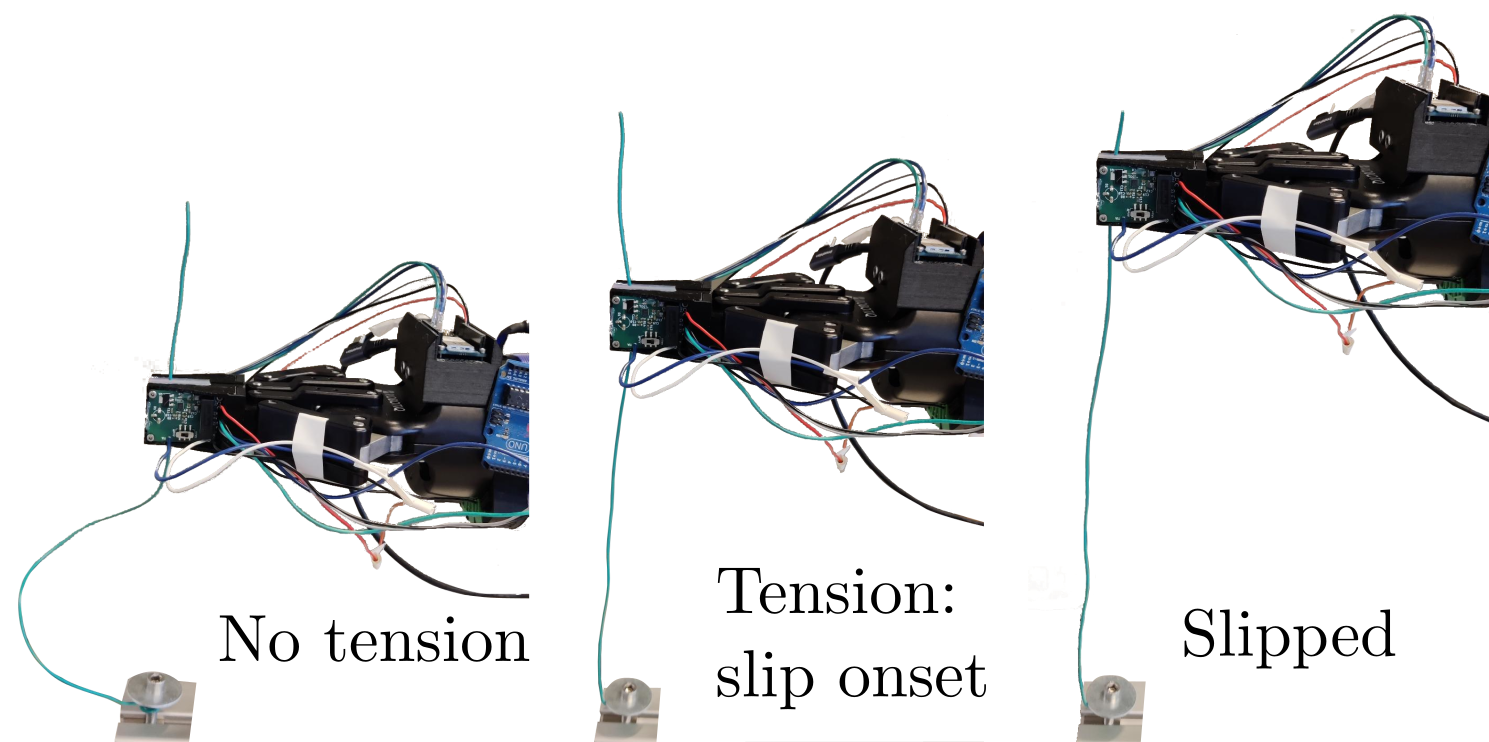} 
    \caption{Setup. The robot moves linearly upward.}
    \label{fig:cable_setup}
\end{subfigure}

\vspace{2mm}

\begin{subfigure}[t]{0.4\textwidth}
    \centering
    \includegraphics[width=0.9\linewidth]{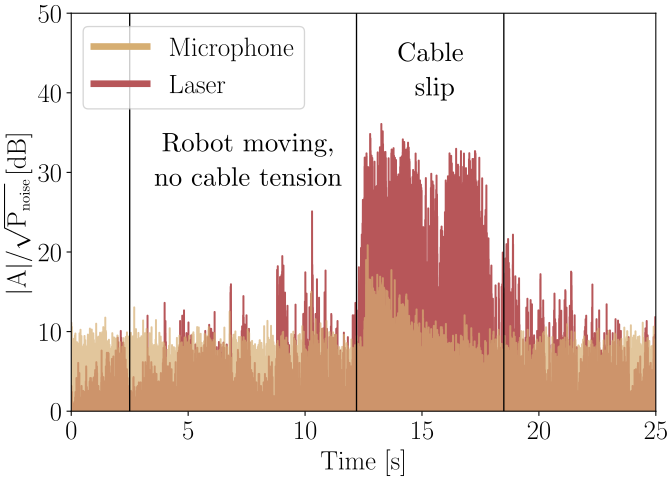} 
    \caption{Normalised sensor output amplitudes, robot linear speed of 1\,cm/s.}
    \label{fig:cable_plot}
\end{subfigure}
\vspace{2mm}

\begin{subfigure}[t]{0.4\textwidth}
    \centering
    \includegraphics[width=0.9\linewidth]{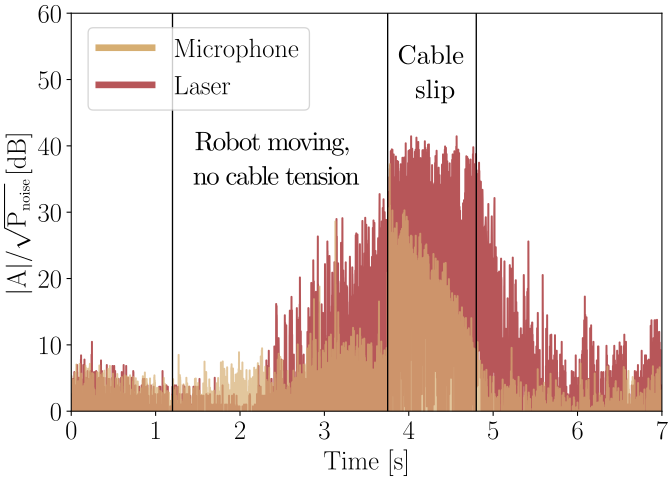} 
    \caption{Normalised sensor output amplitudes, robot linear speed of 5\,cm/s.}
    \label{fig:cable_plot_fast}
\end{subfigure}
\label{fig:cable}
\caption{Measuring cable slip.}
\end{figure}

Fig.~\ref{fig:cable_setup} shows the setup for the first experiment: a \diameter 1.57\,mm electrical wire is attached to the table and grasped by the sensorised gripper.
The gripper moves linearly upward.
Once the cable is in tension, it will start to slip between the fingertips.
This is repeated for movement speeds of 1\,cm/s and 5\,cm/s.

Fig.~\ref{fig:cable_plot} shows the normalised sensor outputs for a speed of 1\,cm/s.
The laser clearly responds more to the slip event: the SNR during slip is 22.2\,dB for the laser and 1.7\,dB for the microphone.
Hence, the laser is 20.5\,dB more sensitive than the microphone in this experiment.
When the movement speed is increased to 5\,cm/s, see Fig.~\ref{fig:cable_plot_fast}, the laser SNR is 30.4\,dB, the microphone SNR is 13.9\,dB, indicating that the laser is 16.5\,dB more sensitive.
The rise of the signal amplitudes before cable slip is attributed to slight movements of the cable being pulled up: if it were motor noise, it would have started earlier.

\subsection{Box slip}
\begin{figure}[tpb]
\centering
\begin{subfigure}[t]{0.45\textwidth}
    \centering
    \vspace{2mm}
    \includegraphics[height=3cm]{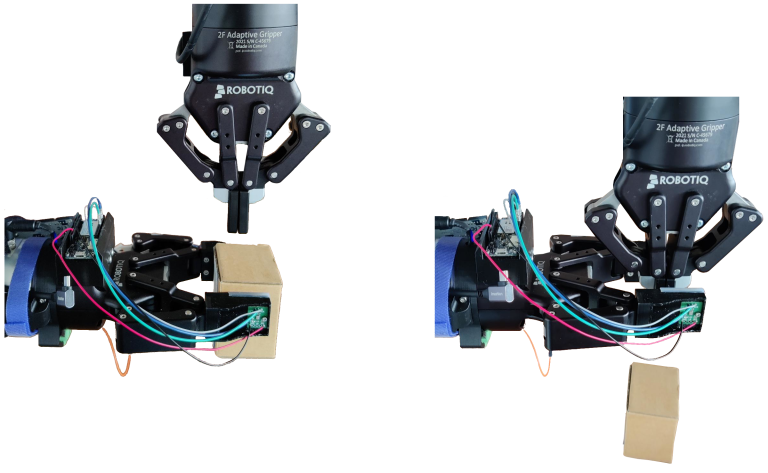} 
    \caption{Setup. The right robot moves down and pushes the box out from between the fingertips of the left robot.}
    \label{fig:box_setup}
\end{subfigure}
\vspace{2mm}

\begin{subfigure}[t]{0.4\textwidth}
    \centering
    \includegraphics[width=0.9\linewidth]{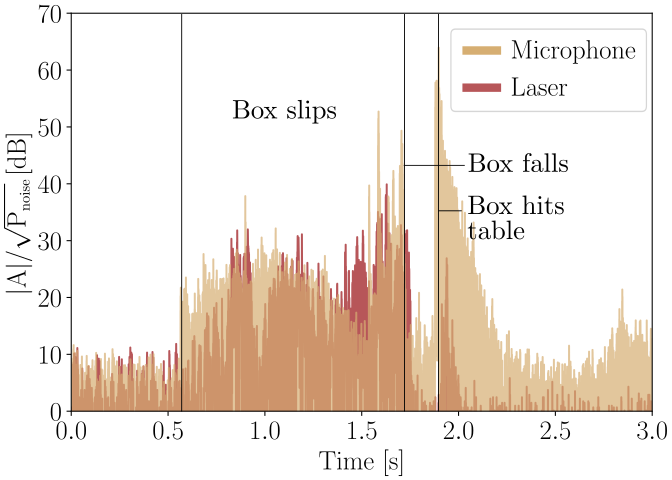} 
    \caption{Normalised sensor output amplitudes, robot linear speed of 2\,cm/s.}
    \label{fig:box_plot}
\end{subfigure}
\vspace{2mm}

\begin{subfigure}[t]{0.4\textwidth}
    \centering
    \includegraphics[width=0.9\linewidth]{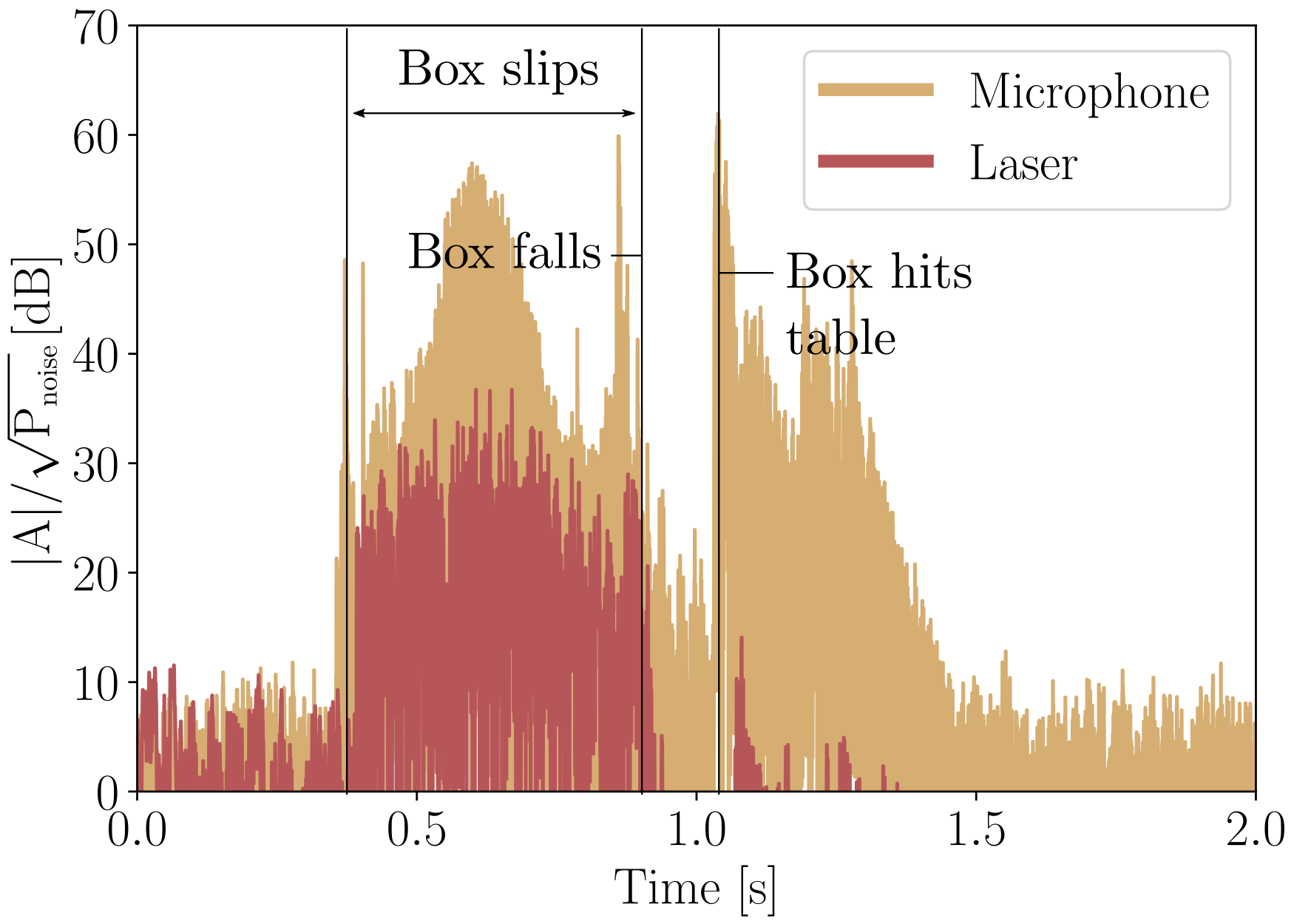} 
    \caption{Normalised sensor output amplitudes. robot linear speed of 5\,cm/s.}
    \label{fig:box_plot_fast}
\end{subfigure}
\label{fig:box}
\caption{Measuring cardboard box slip.}
\end{figure}

For the second slip experiment, a larger object is chosen: a 260\,g cardboard box, see Fig.~\ref{fig:box_setup}. 
The box is held by the sensorised gripper, and pushed from between its fingertips by a second robot.
Again, two different slip speeds are evaluated: Fig.~\ref{fig:box_plot} shows the normalised sensor outputs when the second robot moves at a linear speed of 2\,cm/s, for Fig.~\ref{fig:box_plot_fast}, the robot moves at a speed of 5\,cm/s.

At a slip speed of 2\,cm/s, the SNR is 21.2\,dB for the laser, 25.9\,dB for the microphone.
Notably, the signal peak when contact with the box is lost is about 18\,dB larger for the microphone than for the laser.
When the slip speed is increased to 5\,cm/s, Fig.~\ref{fig:box_plot_fast}, the microphone clearly outmatches the laser with an SNR of 41.8\,dB, 17.7\,dB larger than the laser SNR of 24.1\,dB.

\subsection{Pencil extrinsic contact}

%After tackling tasks like in-hand pose estimation, 
Researchers have recently turned their interest towards extrinsic contact estimation \cite{liu2024maniwav, kim2023, ota2024}.
In ManiWAV\,\cite{liu2024maniwav}, the sound emanating from extrinsic contacts was effectively used to enhance robot learning of complex tasks such as flipping a bagel with a spatula.
Here, we conduct an experiment to gauge whether or not robot learning similar to \cite{liu2024maniwav} could be feasible with SMI sensing.

\begin{figure}[tpb]
\centering
%\vspace{2.7mm}
\begin{subfigure}[t]{0.4\textwidth}
    \centering
    \includegraphics[height=3cm]{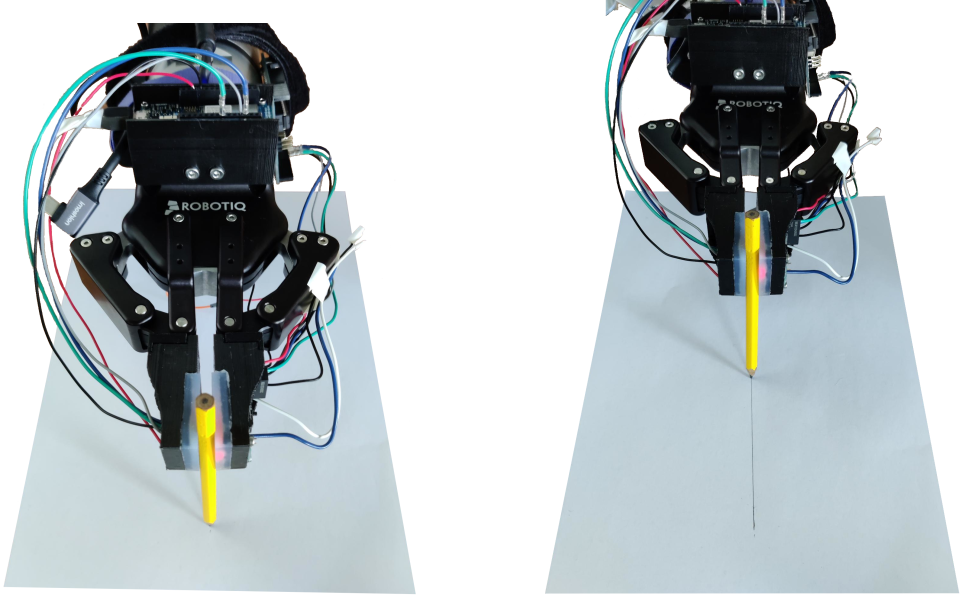} 
    \caption{Setup. The robot moves linearly backwards at a speed of 5\,cm/s while holding the pencil.}
    \label{fig:pencil_setup}
\end{subfigure}

\vspace{2mm}

\begin{subfigure}[t]{0.23\textwidth}
    \centering
    \includegraphics[width=1\linewidth]{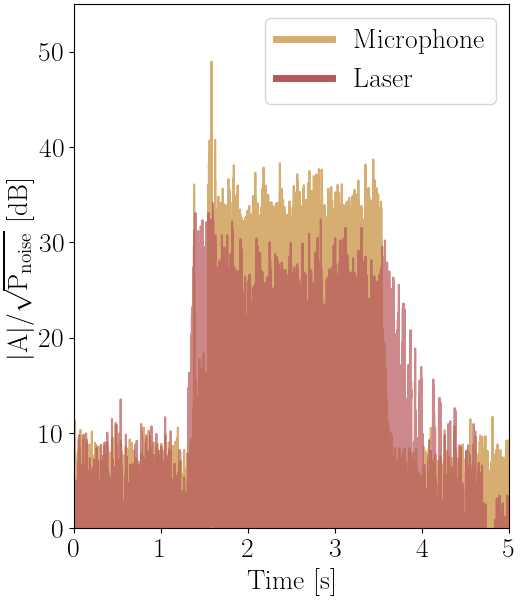} 
    \caption{Normalised sensor output amplitudes, 57\,dB ANL.}
    \label{fig:pencil_plot_noNoise}
\end{subfigure}\hfill
\begin{subfigure}[t]{0.23\textwidth}
    \centering
    \includegraphics[width=1\linewidth]{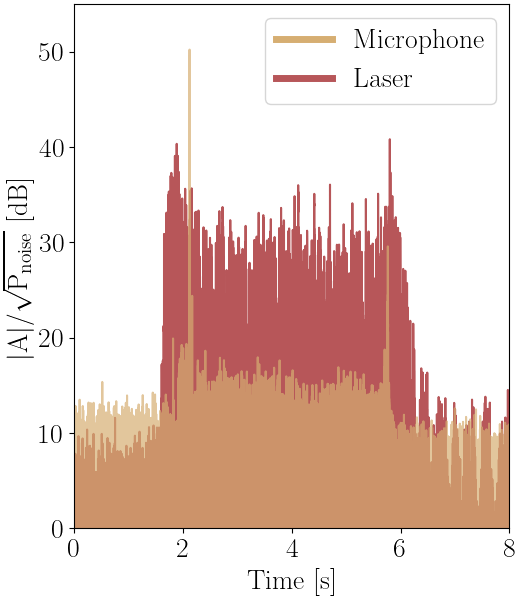} 
    \caption{Normalised sensor output amplitudes, 62\,dB ANL.}
    \label{fig:pencil_plot_noise}
\end{subfigure}
\label{fig:pencil}
\caption{Measuring extrinsic contact of a pencil.}
\end{figure}

Fig.~\ref{fig:pencil_setup} shows the setup: the gripper holds a pencil and moves linearly backwards at a speed of 5\,cm/s.
The sensor outputs are shown in Fig.~\ref{fig:pencil_plot_noNoise}.
The large peaks in the plot are due to sudden starting and stopping of the robot and are not considered for the power calculations.
The SNR of the laser is 19.9\,dB, that of the microphone is 24.5\,dB, a difference of 4.6\,dB.
There is, however, a limitation to microphones that we have not explored yet: their tendency to pick up ambient noise.
When we repeat the experiment with white noise playing from laptop speakers, see Fig.~\ref{fig:pencil_plot_noise}, the microphone performs much worse than the laser: the SNR of the laser is 21.3\,dB, that of the microphone is 5.0\,dB, meaning that the laser picks up the extrinsic contact 16.3\,dB better than the microphone.
The ANL without white noise was measured to be 57\,dB around the fingertips, which increased to 62\,dB while playing the white noise.

\subsection{Cup extrinsic contact}

\begin{figure}[tpb]
\centering
\begin{subfigure}[t]{0.23\textwidth}
    \centering
    \includegraphics[width=1\linewidth]{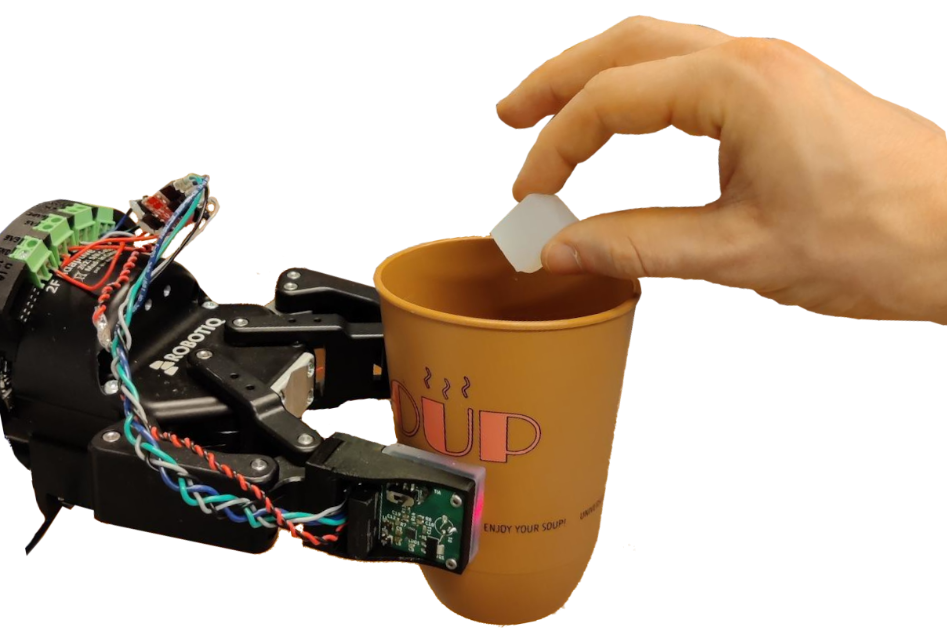} 
    \caption{Pieces of soft silicone are dropped into the cup.}
    \label{fig:cup_silicone_setup}
\end{subfigure}\hfill
\begin{subfigure}[t]{0.23\textwidth}
    \centering
    \includegraphics[width=1\linewidth]{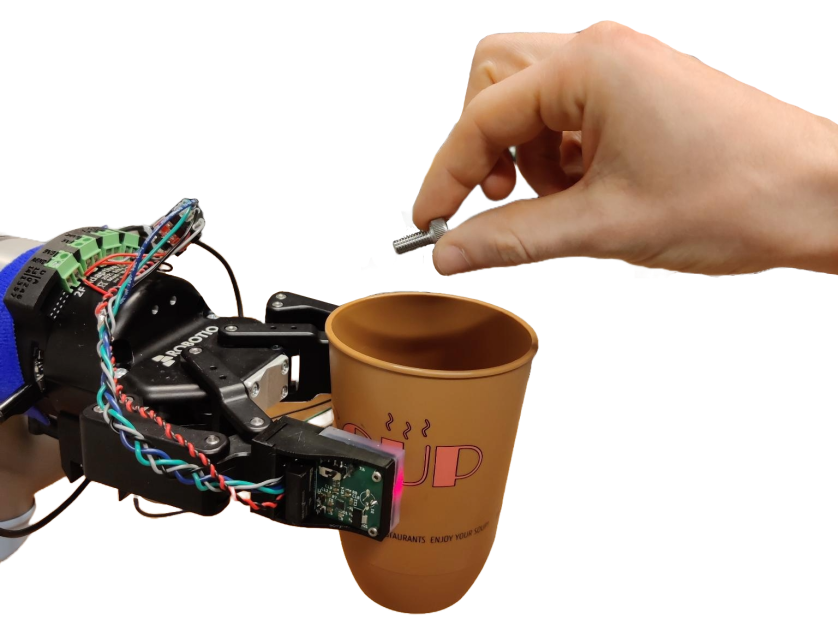} 
    \caption{M6x14 bolts are dropped into the cup.}
    \label{fig:cup_bolt_setup}
\end{subfigure}

\vspace{2mm}

\begin{subfigure}[t]{0.23\textwidth}
    \centering
    \includegraphics[width=1.05\linewidth]{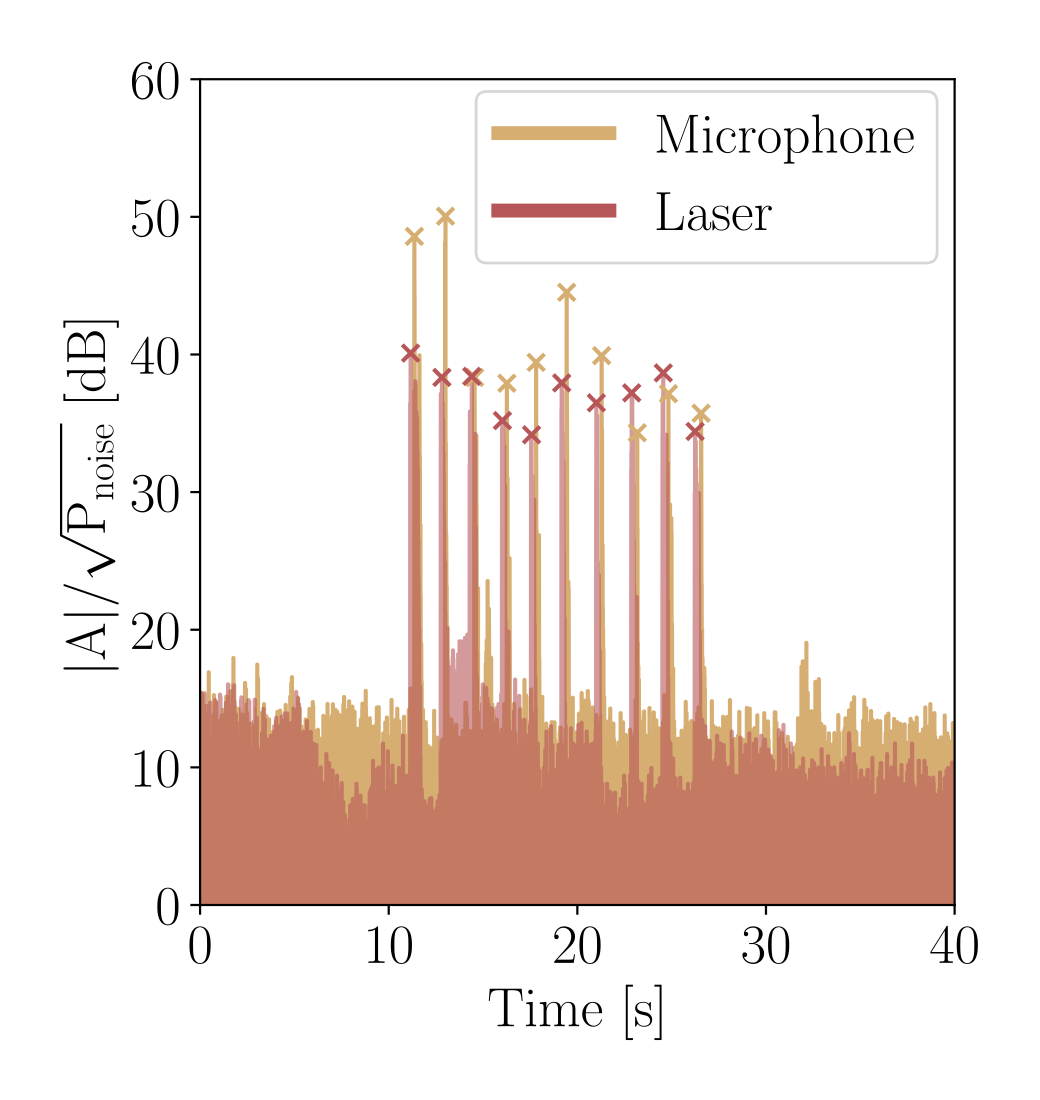} 
    \caption{Silicone: normalised sensor output amplitudes, 57\,dB ANL.}
    \label{fig:silicone_plot}
\end{subfigure}\hfill
\begin{subfigure}[t]{0.23\textwidth}
    \centering
    \includegraphics[width=1.05\linewidth]{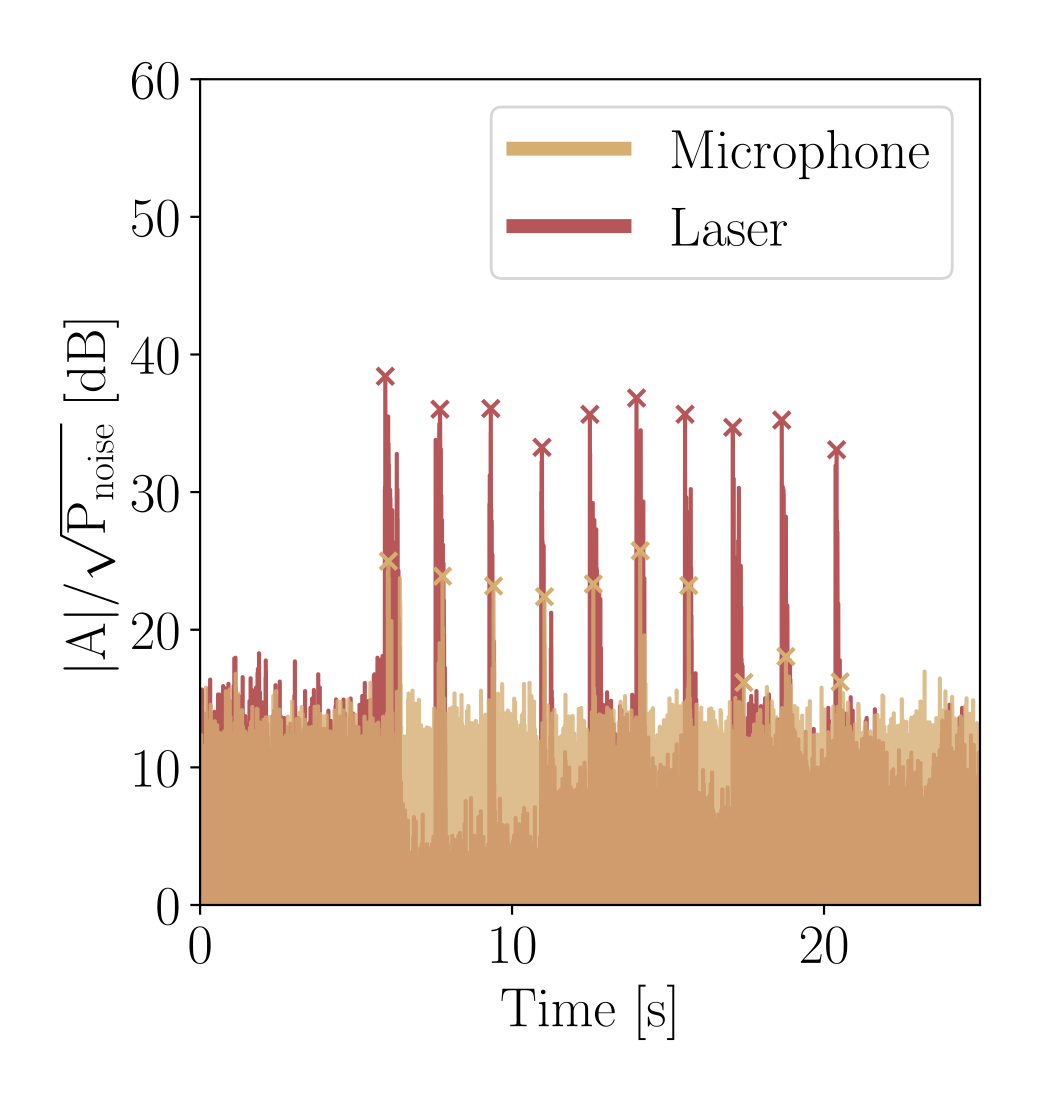} 
    \caption{Silicone: normalised sensor output amplitudes, 82\,dB ANL.}
    \label{fig:silicone_noise_plot}
\end{subfigure}

\vspace{2mm}

\begin{subfigure}[t]{0.23\textwidth}
    \centering
    \includegraphics[width=1.05\linewidth]{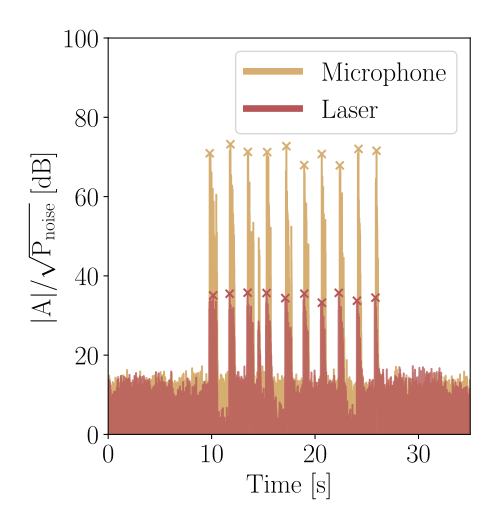} 
    \caption{Bolts: normalised sensor output amplitudes, 57\,dB ANL.}
    \label{fig:bolts_plot}
\end{subfigure}\hfill
\begin{subfigure}[t]{0.23\textwidth}
    \centering
    \includegraphics[width=1.05\linewidth]{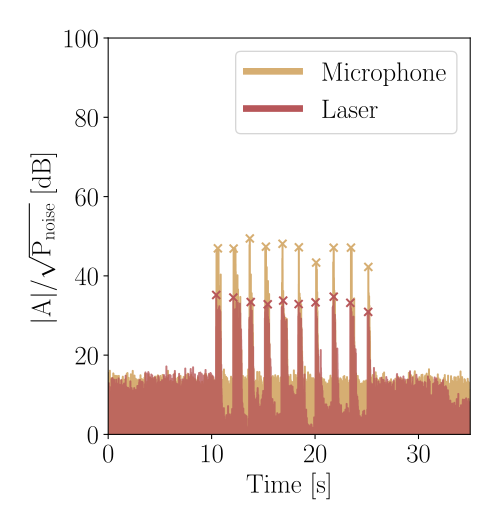} 
    \caption{Bolts: normalised sensor output amplitudes, 82\,dB ANL.}
    \label{fig:bolts_noise_plot}
\end{subfigure}
\caption{Measuring extrinsic contact of objects falling into a hard plastic cup.}
\label{fig:cup}
\end{figure}

Another extrinsic contact experiment in ManiWAV\cite{liu2024maniwav} consists of using audio cues to guess whether a cup is empty or not while shaking it.
We perform an analogous experiment by having the robot hold a hard plastic cup and dropping either ten soft silicone pieces (Silicone Addition Colorless 5 by Silicones and More, 3.6\,g on average) or ten 14\,mm M6 bolts into the cup.
Dropping the objects sequentially rather than having the robot shake the cup provides more consistent signal peaks.
We repeat these experiments both with and without added external white noise.
Here, the SNR is calculated as:
\begin{equation}
    \text{SNR}=\frac{\sum_{i=1}^{10} |p_i|^2}{\text{P}_{\text{noise}}\cdot10}
\end{equation}
with $p_i$ the peak amplitude corresponding to the $i$-th dropped object.
The results are shown in Fig.~\ref{fig:cup}.

When silicone pieces are dropped into the cup, Fig.~\ref{fig:silicone_plot}, the SNR of the microphone is 43.9\,dB, outperforming the laser by 6.4\,dB.
The first two peaks in the microphone signal are quite high compared to the others, because the first silicone pieces fall directly into the cup, while the others make a gentler landing on the preceding silicone pieces.
With external white noise (Fig.~\ref{fig:silicone_noise_plot}), the laser outperforms the microphone by 13.0\,dB.
The laser SNR only dropped 1.7\,dB when 25\,dB of noise was added, the microphone SNR dropped 21.1\,dB.
When dropping bolts, Fig.~\ref{fig:bolts_plot}\nobreakdash--\ref{fig:bolts_noise_plot}, even at 82\,dB ANL, the microphone SNR is 6.4\,dB better than the laser.

\section{Discussion}

\subsection{Design}

The ADL-65075TL is one of the cheapest lasers available, costing around \euro{4}.
The fact that no high-end components or lenses are needed makes for a scalable design.
The required filters and amplifiers (Fig.~\ref{fig:circuit}) do increase the size of the SMI PCB (Fig.~\ref{fig:fingertips} and \ref{fig:section_view}), whereas microphones have already benefited from decades of technological advances and are now ubiquitous fully integrated devices. 
Given the precedent of the integrated LDV developed in \cite{morita2015}, and the fact that SMI circuits are simpler than LDVs, an integrated SMI sensor is feasible.
SMI's characteristic non-contact sensing allows for the electronics to be integrated away from the finger contact surface, making for a robust design. 
Alternatively, other sensors can be included on the finger surface. 
Multimodal sensor integration is a promising way of enhancing the perception capabilities of robots~\cite{wang2023}.

%In addition, \cite{contactile} has demonstrated a 3D force and displacement sensor based on reflection of light within an enclosed cavity.
%Since the highly sensitive SMI modality also floods an air cavity with light, we can envision a system similar to \cite{contactile} to achieve 3D force sensing on top of the sensing capabilities demonstrated in this paper.

\subsection{Robot Manipulation Experiments}

\begin{figure}
\centering
\includegraphics[width=0.88\linewidth]{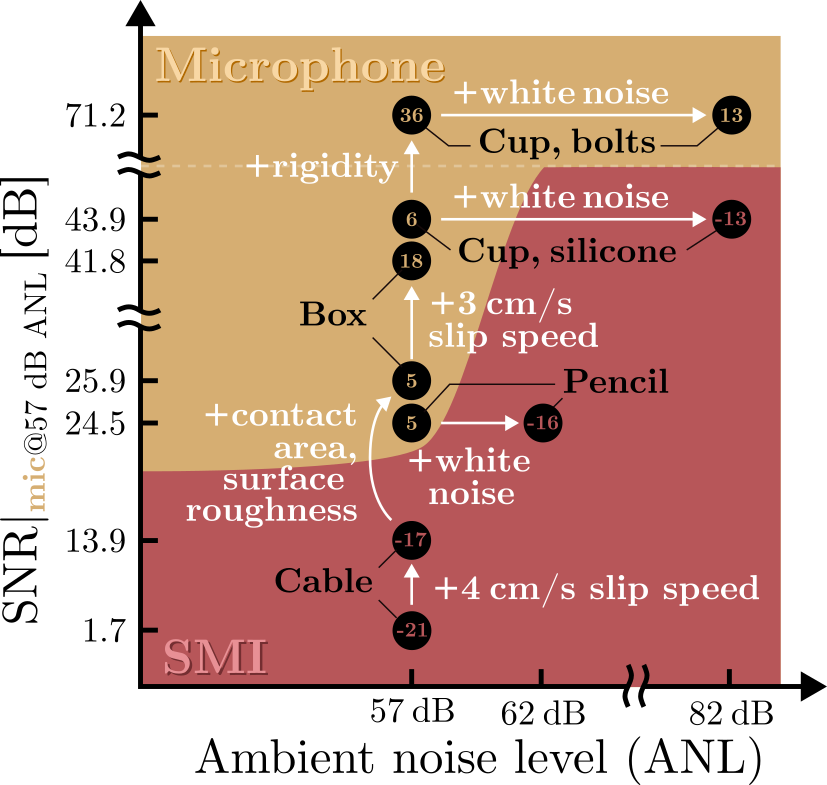}
\caption{Technology decision map. Each experiment is indicated by a black circle. For each experiment, the value inside the circle denotes the difference in SNR between the microphone and the laser.}
\label{fig:classification}
\end{figure}

We have distilled the experiments from section~\ref{s:experiments} into Fig.~\ref{fig:classification}.
The horizontal axis indicates the ambient noise level.
The vertical axis is the SNR of the microphone at baseline ANL, which in our environment corresponds to about 57\,dB.

%The experiments are ordered along the \enquote{Signal level} axis based on the relative signal power of the microphone, such that \enquote{Signal level} relates to the intuitive notion that a larger slip event causes more audible sound.
The cable slip experiment showed that SMI sensing is able to detect more subtle slip events than the microphone.
The cardboard box, on the other hand, was more easily detected by the microphone while slipping.
We attribute this to the larger contact surface and higher roughness of the cardboard compared to the electrical wire.
However, at a slip speed of 2\,cm/s, the difference was small: 4.7\,dB (a power factor of 3), compared to the cable slip experiment where the difference ranged from 16.5\,dB (power factor 45) to 20.5\,dB (power factor 112).
At a slip speed of 5\,cm/s, the microphone clearly outperformed the laser.
The slip experiments are consistent: a higher slip speed makes the performance difference between both sensors (Fig.~\ref{fig:classification}, values in black circles) shift in favour of the microphone.
Next, the pencil experiment showed that detecting extrinsic contact is feasible with both technologies.
At a movement speed of 5\,cm/s, the difference of 4.6\,dB was again comparatively small.
When adding broadband ambient noise, SMI sensing showed remarkable resilience, whereas the microphone was severely affected.
The latter conclusion was confirmed by the cup experiment.
However, dropping rigid bolts instead of soft silicone into the cup showed that, in some cases, no reasonable amount of ambient noise will make the laser outperform the microphone.

The general intuition is clear: SMI can offer increased sensitivity for the most subtle slip events and in the presence of broadband ambient noise, though microphones still reign for tasks where manipulations cause clearly audible sounds. 
Overall, the SNR difference is a strong indicator of how large the change in ANL should be for one modality to outperform the other.
In addition, when considering the pencil, box, and cable experiments, the location of the boundary between SMI and microphone is consistent: rising microphone SNR (vertical axis in Fig.~\ref{fig:classification}) correlates with a rise in the SNR difference between both sensors (values in black circles).
However, this consistency is broken by the cup experiment: with the silicone pieces, the microphone SNR was high, but the SNR difference between sensors rather small.
However, it may not be appropriate to compare the impulse-like signals of the cup experiment directly with the longer-duration signals of the others; more experiments of both types are required. 

A spectral analysis may prove SMI to be even more effective in the cases where our time domain analysis gave the edge to acoustic sensing.
This is because the nature of SMI places clear limits on the signal amplitude: as soon as the vibration amplitude exceeds $\lambda/2$, larger vibrations should only induce higher frequencies in the signal, rather than higher amplitudes.
%Hence, spectral analysis may push the decision boundary in Fig.~\ref{fig:classification} up along the \enquote{Signal level} axis, in favour of SMI.
Lastly, it should be noted that SMI is limited compared to the current literature on slip detection~\cite{romeo2020}, because only gross slip without a notion of slip direction can be directly inferred.
However, its potential application in extrinsic contact detection (like in ManiWAV~\cite{liu2024maniwav}) with added resilience to broadband ambient noise makes for an exciting research direction.

%Secondly, SMI is highly robust against ambient noise.
%This is an attractive property for robot learning based on extrinsic contact like~\cite{liu2024maniwav}.
%Of course, design effort could be made to improve microphone performance by better ambient sound isolation, but this is non-trivial and our comparison is fair because the fingertip structures are so similar (Fig.~\ref{fig:mechanical}): the benefit of SMI is that it does not require such sound proofing.

\section{Conclusion and Future Work}

We have developed and validated the first-ever robotic fingertip capable of SMI sensing.
We chose to compare SMI sensing with acoustic sensing because of two major similarities: (1) they both sense microvibrations, (2) without requiring mechanical contact with their target.
The latter means that the electronics can be integrated away from the contact surface, making for a robust design. 
Alternatively, multimodal integration with sensors on the surface is possible.
Through several experiments, it was found that SMI can sense more subtle slip events than acoustic sensing and that SMI sensing is significantly more resilient against ambient noise.
In future work, a spectral analysis at high sampling frequency should be conducted to further investigate the sensing capabilities of SMI. 
Next, we plan to employ our SMI fingertip to demonstrate learning of contact-rich tasks, confirming the results of ManiWAV~\cite{liu2024maniwav} and evaluating to what extent ambient noise resilience can ease learning.

\balance
\bibliographystyle{IEEEtran}
\bibliography{references.bib}

\end{document}